\documentclass{sig-alternate}
\usepackage{float}
\usepackage{graphicx}
\usepackage{enumitem}

\begin{document}

\title{Ensemble Maximum Entropy Classification and Linear Regression for Author Age Prediction}
\numberofauthors{3} 

\author{
\alignauthor
Joey Hong \titlenote{Undergraduate at Caltech Class of 2019. JPL Data Science Intern for Summer 2016.}\\
       \affaddr{California Institute of Technology}\\
        \affaddr{Pasadena, CA 91106} \\
       \email{jhhong@caltech.edu} 
\alignauthor Chris Mattmann \titlenote{JPL Co-mentor.} \\
	\affaddr{Jet Propulsion Laboratory}\\
        \affaddr{La Cañada Flintridge, CA 91011} 
\alignauthor Paul Ramirez \titlenote{JPL Co-mentor.} \\
	\affaddr{Jet Propulsion Laboratory}\\
        \affaddr{La Cañada Flintridge, CA 91011}
}
\date{August 2016}

\maketitle
\begin{abstract}
The evolution of the internet has created an abundance of unstructured data on the web, a significant part of which is textual. The task of author profiling seeks to find the demographics of people solely from their linguistic and content-based features in text. The ability to describe traits of authors clearly has applications in fields such as security and forensics, as well as marketing. Instead of seeing age as just a classification problem, we also frame age as a regression one, but use an ensemble chain method that incorporates the power of both classification and regression to learn the author's exact age. 

\end{abstract}


\section{Introduction}
The rapid expansion of the Web is coupled with a startling increase in communication among members of the world via blogs, emails, forums, and social media platforms. Thus, it has become of increasing interest to analyze the authorship of Internet content, and be able to use this massive data to predict characteristics If a person from their use of language. In the task of author profiling, such characteristics can vary along dimensions of age, gender, and personality type, though we will focus on age, which has conventionally been the most difficult ~\cite{PAN}. To be able to do so with high predictive power will have applications in a wide spectrum of fields, from law enforcement, to focused advertising. For example, such technology could be used to evaluate the truthfulness of self-reported ages in forums, and detect underage and illegal trading of products.

In this paper, we introduce chained heuristic solution to author profiling, using both classification and regression to predict author age. As in any Machine Learning algorithm involving text, we first extract relevant textual features and find a suitable representation of documents. We use Natural Language Processing techniques to extract content-based and stylistic features from text, and vectorize the frequencies of such aggregated features for each document. 

Our next step is using the vectorized documents in a Maximum Entropy Classifier, which has proven to be a viable and competitive algorithm in the NLP domain. Author age profiling becomes a problem of multi-class classification into age groups: xx-17|18-24|25-34|35-49|50-64|65-xx. We will analyze the effects of different feature vectors on the accuracy of MaxEnt classification in the Section 4.

Using the predicted age category, we transform the classification problem by using linear regression to map age to a continuous variable rather than a categorical one. Our regression implementation is simple LASSO regression using the same document vectors, but we add the predicted category from classification as another feature, so that the result from classification also influences what age is predicted. In Section 4, we will compare the performance of our chained method to that of ordinary linear regresssion, as outlined in Nguyen et al. ~\cite{Nguyen}. 

The paper will proceed as follows. A survey of related work is presented in Section 2. A mathematical formulation of the MaxEnt model and other theoretical techniques is in Section 3. Section 4,5 overviews the datasets used and the feature extraction process. In Section 6 we will explain in detail our methodology and implementation, and we will discuss the results of our experiment in Section 7,8. Finally, in Section 9 we draw conclusions and present future plans for extending the age prediction task. 

\section{Related Work}
There has been considerable work recently in the task of author profiling. In feature selection, Houvardas and Stamatatos in 2006 showed extracting N-grams as an effective feature for classifying in the domain of author profiling ~\cite{Houvardas}.  In 2007, Estival et al. experimented with linguistic feature selection in author profiling, and revealed POS frequencies as a viable feature ~\cite{Estival}. Calix et al. performed author matching from emails using 55 different stylistic features, along with a K-nearest Neighbors implementation, to achieve $76.72\%$ accuracy ~\cite{Calix}. 

Specifically on age identification, Pennebaker and Stone used LIWC to find a relation between the use of language as a person ages ~\cite{Pennebaker}. The problem has, however, been almost exclusively modeled as a classification problem, with a variety of machine learning methods ranging from Bayesian Methods, to Support Vector Machines and Random Forests. In 2011, Nguyen et al. were the first to treat age as a continuous variable via linear regression, and experimented with regression on various individual corpora ~\cite{Nguyen}.  As far as we are aware, though, we are the first to attempt a chained ensemble method of classification and regression to predict exact age.

\section{Maximum Entropy Classifiers}
Maximum Entropy classification is frequently used in NLP tasks such as POS tagging and sentence boundary segmentation. Due to its success with textual features, we decided to implement a MaxEnt classifier to perform age categorization. We will derive the foundations of the MaxEnt model in NLP, as well as the Generalized Iterative Scaling algorithm employed by Apache OpenNLP to solve the model.  

We let $\widetilde{p}(c, d)$ denote the empirical probability of a training sample (c, d) with document vector $d$ and category $c$, and $N$ be the size of the training set. We also introduce a function for each feature $f_{i,j}$ in our document vector representation, such that:
$$ f_{i, j}(c, d) =  
\begin{cases}
\dfrac{N(w_i)}{\sum_{l}N(w_l)}, \quad \text{if $c = c_{j}$, and $d$ contains $w_i$}, \\
0, \quad \text{otherwise}.
\end{cases}
$$
where $c_j$ denotes the possible categories, and $w_{i}$ the possible context tokens, and $N()$ is a real valued function for counts within the document. Apache OpenNLP uses binarized document vectors by default, though we evidently edited it to use normalized frequency counts. 

Since each feature $f_{i,j}$ is a real valued function, it must have an expected value, calculated empirically by the training data:
$$ \widetilde{p}(f_{i,j}) = \sum_{c,d} \widetilde{p}(c, d) f_{i,j}(c,d).$$
We can also calculate the actual expected value $p(f_{i,j})$ from conditional probability:
$$p(f_{i,j}) = \sum_{c,d} p(d) p(c\,| \,d) f_{i,j}(c,d). $$
We look to restrain out MaxEnt model's conditional distribution to satisfy the empirical value, so that, 
\begin{align*}
\sum_{c,d} p(d) p(c\,| \,d) f_{i,j}(c,d) &= \frac{1}{N} \sum_{c,d} p(c\,| \,d) f_{i,j}(c,d) \\
= \sum_{c,d} \widetilde{p}(c, d) f_{i,j}(c,d) &= \frac{1}{N} \sum_{c,d} f_{i,j}(c,d),
\end{align*}
where we approximate $p(d) = \widetilde{p}(d) = 1/N$.

Thus, we want to find a conditional distribution for our model $p$ which satisfies:
\begin{enumerate}
\item $p(c \, | \, d) \geq 0 \quad \forall c,d$.
\item $\sum_{c} p(c \, | \, d) = 1 \quad \forall d$.
\item $\sum_{c,d} p(c\,| \,d) f_{i,j}(c,d) = \sum_{c,d} f_{i,j}(c,d) \quad \forall i,j$.
\end{enumerate}
From information theory, we know that the optimal model $p^*$ which satisfies the above constraints, is as close to uniform as possible, and is achieved by maximizing the entropy,
$$ p^* = \max_{p} \Big( - \sum_{c,d} p(c\,| \,d) \log{p(c\,| \,d)}\Big). $$

If we set up and solve the Lagrangian ~\cite{Berger}, we get the multi-class version of the sigmoid function in logistic regression as our optimal solution, 
$$ p^*(c \, | \, d) = \dfrac{\text{exp}(\sum_{i,j} \lambda_{i,j} f_{i,j}(c,d))}{\sum_{c} \text{exp}(\sum_{i,j} \lambda_{i,j} f_{i,j}(c,d))}. $$

\subsection{Generalized Iterative Scaling}
Estimating the optimal $\lambda$ parameters can be easily done with an iterative scaling algorithm. OpenNLP employs the simple GIS algorithm to do so. 

It is guaranteed that no local maxima exist in the likelihood surface, so  hillclimbing algorithms from any random initial distribution will find the global maximum entropy solution. We use the log likelihood as with Della Pietra et al. in 1997 ~\cite{Berger}. Let $\lambda$ be the parameter vector of all $\lambda_{i,j}$ values, and $p_{\lambda}$ be the resulting probability model. The log likelihood is given by,

\begin{align*}
\ell(\lambda) &= \log \prod_{d} p_{\lambda} (c \, | d) \\
&= \sum_{d}\sum_{i,j} \lambda_{i,j} f_{i,j}(c,d) \\
&\quad - \sum_{d} \log \sum_{c} \text{exp}\big(\sum_{i,j} \lambda_{i,j} f_{i,j}(c,d)\big).
\end{align*}

As in any hillclimbing algorithm, we seek $\Delta\lambda$ such that $\ell(\lambda + \Delta\lambda) - \ell(\lambda) > 0$. From Jensen's inequality, we can lower bound the difference by an auxiliary function $B$ ~\cite{Nigam},

\begin{align*}
\ell(\lambda + \Delta\lambda) &- \ell(\lambda) \geq B = \\
&1 + \sum_{d} \sum_{i,j} \Delta\lambda_{i,j}f_{i,j}(c, d) \\
& - \sum_{c} p_{\lambda} (c \, | \, d)\text{exp}\Big(f^{\#}(c,d)  \Delta\lambda_{i,j} \sum_{i,j} \frac{f_{i,j}(c,d)}{f^{\#}(c, d)}\Big),
\end{align*}
where $f^{\#}(c,d) = \sum_{i,j}f_{i,j}(c,d)$ is the sum over all features. 

It becomes easy to find $\Delta\lambda$ by merely differentiating $B$ with respect to $\Delta\lambda_{i,j}$, and setting it equal to zero to maximize $B$. 
\begin{align*}
\frac{\partial B}{\partial \Delta\lambda_{i,j}} &= \sum_{d} f_{i,j}(c, d) \\
&\quad - \sum_{c} p_{\lambda} (c \, | \, d) f_{i,j}(c,d) \text{exp}(\Delta\lambda_{i,j}f^{\#}(c, d)).
\end{align*}
So, the GIS algorithm effectively solves for $\Delta\lambda_{i,j}$ and increments for each parameter $\lambda_{i,j}$. 

\medskip
\noindent\rule{8cm}{0.4pt} \\
\begin{enumerate} [nolistsep]
\item \textbf{Input:} Set of labelled documents $d \in \mathcal{D}$ and feature functions $f_{i,j}$.  
\item Initialize $\lambda_{i,j} = 0$ for all $i, j$. 
\item Repeat until convergence:
\item \quad For each $\lambda_{i,j}$:
\item \quad\quad Let $\Delta\lambda_{i,j} = \dfrac{1}{f^{\#}(c, d)} \log \dfrac{\widetilde{p}(f_{i,j})}{p(f_{i,j})}$.
\item\quad\quad Set $\lambda_{i,j} \leftarrow  \lambda_{i,j} + \Delta\lambda_{i,j}$.
\end{enumerate}
\noindent\rule{8cm}{0.4pt}
\medskip

\noindent Thus, the resulting $\lambda$ parameters make up our MaxEnt classifier, as desired.

\section {Data}
\begin{figure}
\centering
\includegraphics[width=8cm]{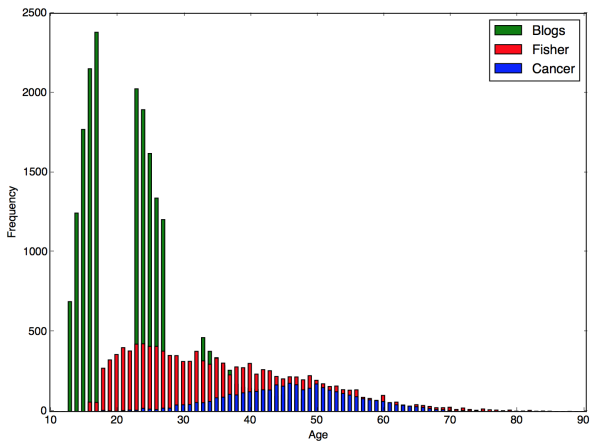}
\caption{Age Frequency Comparison of Authors within Selected Corpera}
\end{figure}

We use labelled data from four major datasets, which were divided via a 90-10 split into a training and test set. Figure 1 outlines the distribution of the three datasets with exact ages, and Table 1 shows the distribution of the PAN16 dataset, with only categorized ages. 

Posts were not counted for the first three corpora since all posts from a single user were aggregated and may vary widely in post length, whereas the PAN16 dataset, which contains solely Twitter data, has relatively uniform post lengths.

\subsection{Blog Authorship Corpus}
The corpus was the result of crawled blogs (blogger.com) by Schler et al. in 2004 ~\cite{Schler}. The corpus consists of posts by 19,320 bloggers, and 681,288 posts, labeled with blogger id, and self-provided gender and age, as well as industry and/or astrological sign when available.

The blogs were formatted in an XML, and we extracted all the text in the files belonging to a user, and aggregated them to a single labelled post.

\subsection{Fisher English Transcripts}
The Fisher CALLHOME dataset contains transcripts of telephone conversations between random pairs of people, where a truth file was provided with characteristics such as age and gender of each speaker ~\cite{Fisher}. Each telephone conversation was regarding a pre-specified topic, which greatly varied over all the recorded conversations. 

We parsed the truth file, and used a script to aggregate all the text among the transcripts belonging to the individual into a single sample. 

\subsection{Cancer Forum}
Posts and user profiles from an online forum for persons with breast cancer (community.breastcancer.org) were crawled by Nguyen et al. in 2011 ~\cite{Nguyen}. The age of the users were either ascertained from context in their posts, or manually annotated by looking at their profiles. Posts from a total of 1,996 users were recorded. 

Again, we aggregated all posts from a single user into one labelled post per user.

\subsection{PAN Dataset}
The PAN16 dataset is comprised of tweets from twitter users, along with an annotated age and gender truth set. The Twitter corpus was compiled by looking at LinkedIn profiles, and finding ones with a corresponding Twitter account. Age was determined either by a provided birth date in the profile, or via an estimation with the degree starting date in the education section ~\cite{PAN}. 

Due to Twitter terms of service, only the URLs to each profile's tweets were provided, and a script was used to download the tweets, and another was written to match downloaded tweets with the provided truth set. 

Ages are labelled with categories: 18-24|25-34|35-49|50-64|65-xx. Because of this, the PAN16 dataset was only used in the training data for our classification task.

\begin{table}
\centering
\begin{tabular}{|| c | p{1.5cm} p{1.5cm} ||} \hline
Age Category & Users & Posts \\ \hline \hline
18-24 & 26 & 29382 \\ 
24-34 & 136 & 41907 \\ 
35-49 & 182 & 44532 \\ 
50-64 & 78 & 35603 \\
65-xx & 6 & 1350 \\ \hline
Total & 428 & 152774 \\
\hline \end{tabular}
\caption{Description of the PAN16 Dataset}
\end{table}

\section{Features}

\subsection{Content-based Features}
It can be assumed that since people of different ages will speak about different topics, the usage of words between ages can be a distinguishing feature in their written text. Since content can be represented as an N-gram feature, we compiled unigrams and bigrams from each set. For unigrams, we put all the words through a list of 319 English stopwords, and removed all occurrences of stopwords, which provide no meaningful content. 

For both classification and regression, for each of the features, we calculated the frequency of appearance in the document, and in creating a vector for each document, used the p-normalized counts in $L^{1}$ space.

\subsection{Stylistic Features}
Style based features include sentence and word counts, punctuation counts, and distribution of Parts of Speech unigrams and bigrams in the document. All of the stylometry was done via Apache OpenNLP, where a custom tokenizer was written to tokenize on whitespace and punctuation, which were consequently fed into OpenNLP's POS tagger for a POS count.

We also counted occurrences of words that belong to LIWC word classes ~\cite{LIWC}, including personal words (``I", ``me", ``mine", etc.), positive (``cheery", ``bliss", ``joy", etc.) and negative (``abandoned", ``bad", ``hurt", etc.) sentiment words, and quantifier words (``many", ``few", ``both", etc.). A total of word lists for $9$ word classes were compiled manually. 

While originally, lemmatization and stemming were employed on all words in the document, in order to preserve stylistic abnormalities, we did not use those tools in the tokenization process. Like with content-based features, the document vectors used normalized counts. 

\subsection{Normalization}
To curb the feature count, and thereby keep the training time of the model in a feasible range, we implemented a Chi-Squared Selector to work alongside OpenNLP's data indexer, and transform the input features into a reduced feature space. We employed a one-vs-many transformation, calculating the $\tilde\chi^2$ for each category relative to all other age categories. We kept the critical value at $\tilde\chi^2 = 2.71$, which meant the assumption of independence could be rejected with 90$\%$ confidence. 

\section{Architecture}

Figure 2 outlines the pipeline of the algorithm, which we will describe in detail. Initially, the text data from all the data sources are aggregated by user. Since most datasets were publicly released, no other external work aside from reformatting to match the input requirements of our classifier and regression trainer was needed. 

\begin{figure}
\centering
\includegraphics[width=8.5cm, height=12cm]{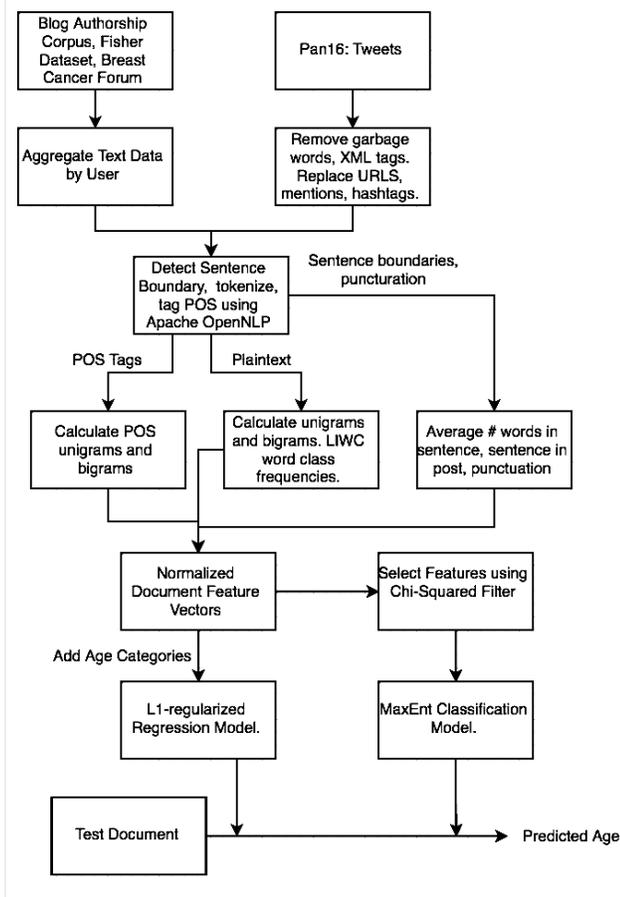}
\caption{System Architecture of Algorithm}
\end{figure}

\subsection {Data Preprocessing}
As stated, most datasets required no significant cleaning. The PAN16 twitter data, however, was raw XML text and required cleaning tags and removing non-encodable words. We used regex filtering to replace all outlinks with ``urllink", and removed all hashtags and mentions from the data. The cleaned data can then be formatted into the same input requirements. 

Since the compiled datasets had a bias towards people in the $20$'s, we used oversampling to duplicate all posts from infrequent age classes and correct for the bias. 

\subsection {Feature Selection}

We used Apache OpenNLP's POS tagging and sentence detector tools to extract stylistic features from the formatted data. Our custom tokenizer separated based on both whitespace and punctuation, and boundaries for sentences. The tokenized text was indexed into a count vectorizer to extract features such as $\#$ of sentences in document, words in sentence, and punctuation classes. 

Using the tokenized text, unigrams and bigrams were extracted.  Unigrams were fed into a stopwords filter, where all tokens matching stopwords in a preset list were removed. We also counted occurrences of unigram words in each of our manually created LIWC word classes, and calculated the distribution of frequencies. 

We then pipelined the tokens into the POS tagger to generate POS unigrams and bigrams, and a frequencies were again extracted with the POS tags. Finally, all feature frequencies were normalized on length of document.  

Feature extraction was done using feature selectors from Apache OpenNLP, and also parallelized on Apache Spark's distributed dataset framework.

\subsection {Classification}

For classification, all the features were used, but filtered via statistical feature selection methods. We only selected features that occurred at least $10$ times among the training documents. Our normalized document vectors were also put through a Chi-Squared feature selector, so only features with 90$\%$ confidence of dependence were selected. 

We leveraged Apache OpenNLP's Generalized Iterative Scaling method to solve the parameters for our MaxEnt classifier model. 

\subsection{Regression}

For regression, we did not use the text bigrams features, as linear regression did not scale well with a high number of features. Also, since the PAN16 tweets were only labeled with age class, we omitted the set from the training data.

We passed in the age categories as another feature in our regression model trainer. Only features that existed in at least 5 different documents were selected. For regularization, we apply L1 regularization to train a model with LASSO, so that our feature weights are optimized but sparse. For tuning the regression parameters, we tested on several learning rates and regularization values ($0.5$, $0.25$, $0.125$, $0.0625$, $0.03125$, $0.015625$). 

To perform our chained author age prediction, we can pass in our classifier model, which first evaluates the top age category, and then adds the category feature into the document vector to be evaluated by our LASSO model. Regression training leveraged L1-regularized regression using stochastic gradient descent from Apache Spark's MLlib.

\section{Experiment}

\subsection{Classification}
We experimented with various feature vectors with the MaxEnt classifier. This is done by training on the datasets using the desired feature generators enabled in our implementation. We get the following models: 
\begin{enumerate}
\item UNIGRAM:  Model trained on the both training corpora with only unigrams.
\item NGRAM: Model trained on joint training corpora with unigrams and bigrams.
\item STYLE: Model trained on joint training corpora with unigrams and stylistic features (sentence, word counts, punctuation, POS, LIWC, etc.)
\item GLOBAL: Model trained on joint training corpora with all above features.
\end{enumerate}

We measured the accuracy of each individual model on a joint test dataset. We also calculated precision, recall, and the F$_{1}$-score for the best-performing model among all the age categories. 

As mentioned earlier, we did a 90-10 split on the entire joint corpora to produce the training and testing datasets. 

\subsection{Regression}
Both regression models were trained using all the features enabled. We have the following two models for regression:
\begin{enumerate}
\item DEFAULT: Model trained without passing in result from classification.
\item ENSEMBLE: Model trained with chained method. We used the GLOBAL classifier in the ensemble. 
\end{enumerate}

We measured the mean absolute error, as well as the Pearson's correlation coefficient, for each model, and generated a report of predicted vs. expected ages to be plotted in a scatterplot. 

\section{Results and Discussions}
\begin{table}
\centering
\begin{tabular}{|| c | p{1.7cm} | p{1.7cm} ||} \hline
Model & \# Features & Accuracy \\ \hline \hline
UNIGRAM  & 60,814 & 0.484  \\
NGRAM  & 391,737 & 0.514 \\
STYLE & 65,464 & 0.489  \\
GLOBAL & 396,390 & 0.521\\
\hline \end{tabular}
\caption{Classification Results for Different Features}
\end{table}

\subsection{Classification}
\begin{table}
\centering
\begin{tabular}{|| c | p{1.5cm} | p{1.5cm} | p{1.5cm} ||} \hline
Category & Count & Precision & Recall \\ \hline \hline
xx-17 & 22217 & 0.701 &  0.537\\
18-24 & 14764 & 0.523 & 0.132  \\
24-34 & 20061 & 0.463 &  0.864 \\
35-49 & 8576 & 0.397 & 0.163 \\
50-64 &  225 &  0.463 & 0.062 \\
65-xx & 50 & 0 & 0 \\
\hline \end{tabular}
\caption{Precision-Recall for GLOBAL Classifier}
\end{table}

As expected, from Table 2, the GLOBAL model with all the features performed best, though only marginally. Only choosing unigrams and bigrams, as in the NGRAM model, performed almost as well in predictive accuracy. 

While the accuracy is not very high, all models performed better than the baseline of $0.33$, calculated by always predicting the most common category, which is xx-17, from the frequencies of age groups among the test data. 

Table 3 highlights the fine-grained report for the best-performing GLOBAL model. It is clear that the model performs increasingly poorly as age group increases, most likely due to the lack of training data among those age groups. 

For the younger age groups, GLOBAL's performance was generally good. For the 18-24 age group, however, many posts in that group were rather predicted to be in the xx-17 age group, resulting in a low recall. This can be because most of the data in the xx-17 age group belonged to people at least 16. Because of this, the algorithm had trouble determining the difference between the two age groups, which could be fixed easily by setting looser boundaries for the age categories, and creating larger age groups as a result. 

Among the older age groups, the algorithm began predicting posts in such categories as xx-17, probably due to the abundance of data in that category, and subsequent lack of data for people older than 50 years old.

\subsection{Regression}

\begin{table}
\centering
\begin{tabular}{|| c | p{2.5cm} | p{2.5cm} ||} \hline
 Model & Correlation (r) & Error (MAE) \\ \hline \hline
DEFAULT  & 0.608 & 12.053 \\
ENSEMBLE  & 0.701 & 10.324 \\
\hline \end{tabular}
\caption{Regression Results}
\end{table}

\begin{figure}
\centering
\includegraphics[width=8cm]{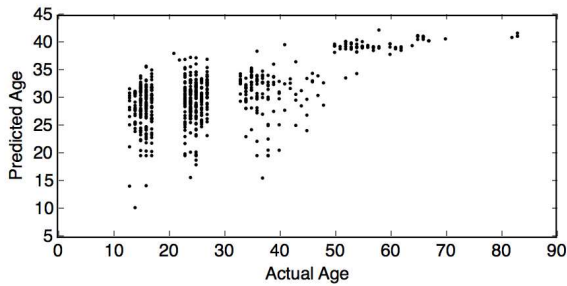}
\caption{ENSEMBLE Scatterplot Age}
\end{figure}

\begin{figure}
\centering
\includegraphics[width=8cm]{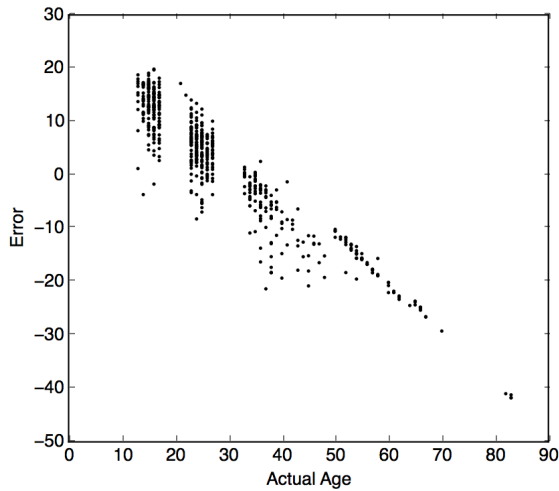}
\caption{ENSEMBLE Scatterplot Error}
\end{figure}

Both regression models suffered from the uneven distribution of data. Though we employed oversampling to mitigate the bias, both models had small coefficients for their features, and predicted relatively close to their respective intercepts. 

While in training, the MAE for both models remained relatively small, the correlation in DEFAULT of $0.608$ gives an $r^2 = 0.37$, which means only $37\%$ of the variance can be explained. In ENSEMBLE, the $r^2$ increases to $0.49$. This is perhaps caused by the bias in training distribution, noisy data, and the fact that most ages lie close to each other, particularly after applying oversampling to the data. 

The MAE of around 10 for ENSEMBLE means it predicts on average, 10 years away from the actual age. While this is high, the mean is skewed by the vast underpredictions for ages older than $50$. It may be that language use does not change much after that age, though the more plausible reason is our lack of training data in the older age range. 

The skew in performance is evident in Figure 3, which plots the predicted vs. actual age for a random sample of 600 in the test data, or in Figure 4, which plots the error for the same sample. 

In addition, the ENSEMBLE model is dependent on the predictions of our MaxEnt classifier, which itself did not have high accuracy. If the classifier predicted age category perfectly, the MAE of the ENSEMBLE model would reduce to around 7. 

\section{Conclusion and Future Work}
We tried studying the relation between language usage and age for the task of author profiling. Our dataset used a collection of annotated corpora, belonging to forum posts, telephone speech, blogs, and tweets, where data from older age groups were duplicated to reduce bias.  
 
We approached the age prediction problem via both classification, where age was separated into age groups, and regression, where age is treated as a continuous variable. 

In classification, we got a predictive accuracy of at best $52\%$, roughly $20\%$ over the baseline, by training a MaxEnt classifier with both content-based and stylistic features. However, precision and recall decreased drastically for older age groups. For the future, we can set better boundaries for users, perhaps by significant life stages rather than arbitrary 10-year age groups. Also, instead of normalized frequencies, we can experiment with tf-idf weights for features. In addition, rather than MaxEnt, we could compare performance to other classification methods, namely Support-Vector Machines or Random Forests.

For regression, we tried both default L1-regularized linear regression and an ensemble method that chains our MaxEnt classifier and passes its predicted age group as a separate category. The performance benefit was marginal, though both models suffered in correlation likely from bias in the training data. The chained regressor had a MAE of around $10$ on the test set, $2$ lower than an MAE of $11$ using default regression. 

Both models predicted well, within an MAE of 6, for ages under 50, but fell drastically in performance for older ages, most likely due to the small coefficients on the features. Future work could involve dimensionality reduction, as SGD does not scale well with many features. Also, by further investigating the coefficients, we can compile a better, more concise list of features.

\nocite{*}

\bibliographystyle{abbrv}
\bibliography{age-prediction}


\end{document}